\documentclass[11pt,a4paper]{article}
\pdfoutput=1
\usepackage[hyperref]{acl2020}
\usepackage{times}
\usepackage{latexsym}

\usepackage{microtype}

\aclfinalcopy

\usepackage{amsmath,amssymb,amsfonts}
\usepackage{graphicx}
\usepackage{xcolor}

\usepackage[utf8]{inputenc}
\usepackage[english]{babel}
\usepackage{xspace}
\usepackage{booktabs}

\usepackage{multirow}
\usepackage{hyperref}
\usepackage{balance}
\usepackage{siunitx}

\newcommand{\ie}{\emph{i.e.,}\xspace}
\newcommand{\eg}{\emph{e.g.,}\xspace}

\newcommand{\albert}{AlbertPT\xspace}
\newcommand{\nome}{BertPT\xspace}
\newcommand{\multi}{Multilingual\xspace}

\title{Mono vs Multilingual Transformer-based Models: \\a Comparison across Several Language Tasks}

\author{
  Diego de Vargas Feijo, Viviane Pereira Moreira \\
  Institute of Informatics, Federal University of Rio Grande do Sul, Porto Alegre, Brazil \\
  \{dvfeijo, viviane\}@inf.ufrgs.br
}
\date{}
    
\begin{document}
\maketitle
\begin{abstract}
BERT (Bidirectional Encoder Representations from Transformers) and ALBERT (A Lite BERT) are methods for pre-training language models which can later be fine-tuned for a variety of Natural Language Understanding tasks.
These methods have been applied to a number of such tasks (mostly in English), achieving results that outperform the state-of-the-art.
In this paper, our contribution is twofold. 
First, we make available our trained BERT and Albert model for Portuguese. 
Second, we compare our monolingual and the standard multilingual models using experiments in semantic textual similarity, recognizing textual entailment, textual category classification, sentiment analysis, offensive comment detection, and fake news detection, to assess the effectiveness of the generated language representations. 
The results suggest that both monolingual and multilingual models are able to achieve state-of-the-art and the advantage of training a single language model, if any, is small.
\end{abstract}


\section{Introduction \label{sec:introduction}}

Text embeddings are powerful tools as they are able to encompass the context, syntactic, and semantics of the texts they represent~\cite{mikolov2013distributed,FastText,glove,Peters:2018}.
Their use has been popularized in the last few years, to a great extent, motivated by the free availability of pre-trained models for a wide number of natural languages. 
These pre-trained language representations can be used as the features that compose the first layer of a task-specific model. Alternatively, one could \emph{fine-tune} a generic model to be applied on a specific task.

More recently, \emph{Bidirectional Encoder Representations from Transformers} (BERT)~\cite{devlin2018bert} and subsequently \emph{A Lite BERT} (ALBERT)~\cite{lan2019albert} were proposed as task-independent architectures.
Both BERT and ALBERT report achieving state-of-the-art results on several natural language processing (NLP) tasks, including next sentence prediction, recognizing textual entailment, sentiment analysis, question answering, and named-entity recognition.
The authors of both models released trained monolingual models in English and Chinese. However, for the other languages, only BERT released a multilingual model that was trained on Wikipedias for 104 languages~\cite{pires2019multilingual}.
Portuguese is among the languages that compose the multilingual model. 
However, it is not clear how the multilingual model performs in single language tasks  compared to a model trained exclusively for that language.

%

We believe that a comparison between the single and multilingual models applied to a variety of single language tasks would help researchers decide whether it is worth to invest training a model on a target language or a multilingual model can be enough.
Also, we make our trained models in Portuguese available for researchers so they could use these models without needing to master the complexity that the training involve.

In this paper, we present \nome and \albert, pre-trained models in Portuguese which are freely available at \url{https://github.com/diego-feijo/bertpt/}.
Both \nome and \albert were trained using corpora coming from different domains and styles (Wikipedia, news articles, movie subtitles, research abstracts, and European Parliament sessions) to assure a wide coverage of the language.

We evaluated our pre-trained models on seven natural language understanding tasks.
Our performance was compared to baselines published in the literature and to the multilingual BERT model.
The results show that \nome and \albert are able to outperform the baselines in most cases.
In relation to the multilingual BERT model, the results of the are within a 5\% proportional differences in most tests.


\section{Related Work \label{related}}

Vector representation of words, known as word embeddings, brought a major advance in NLP.
The features represented by those vectors may help algorithms achieve better performance in a series of tasks.
This happens because those vectors are able to capture several characteristics of the language, allowing algorithms to focus on solving a specific task.
\cite{mikolov2013distributed} presented several examples of the potential of these representations which allow clustering by meaning and vector operations. 

Despite the ability of these vectors to represent several aspects of the language, their fixed representation does not offer flexibility when the meaning of the words needs to be understood within a context. Such a context-independent representation works like a dictionary with a fixed meaning previously determined.
This issue was studied by \cite{Peters:2018} that proposed ELMo (Embeddings from Language Models).
In that approach, the embeddings are a function of all internal layers of a bidirectional language model.

The use of these representations may help tackle the problem of little training data.
This problem happens as the algorithm has to learn both how to solve one specific task and how to represent several aspects of the language.
If a learning algorithm could already understand some behavior of the language, then the specific task to be learned would become easier.
The technique of using the previously learned knowledge and applying it to another model is known as \textit{Transfer Learning}.
There are two common approaches for transfer learning: feature extraction and fine-tuning.
Feature extraction happens when the learned representations from a previous network are used to extract meaningful features from new samples.
%
Fine-tuning~\cite{radford2018improving,devlin2018bert,lan2019albert} adopts a similar approach.
It takes advantage of previously trained weights, but the model is slightly modified.
Typically, the last few layers of the model are replaced by a task-specific layer like a classifier or a linear regression layer.
After that, the entire model is trained, \ie ``fine-tuned'' with a small learning rate, allowing the model to keep part of the learned representations and focus on the specific task.

Another major advance in the last few years was the Transformer~\cite{vaswani2017attention} model. 
It allowed avoiding the need for purely recurrent models. 
Subsequent models like BERT~\cite{devlin2018bert}, Generative Pre-trained Transformer~\cite{radford2018improving}, RoBERTa~\cite{liu2019roberta}, and ALBERT~\cite{lan2019albert} use the Transformer architecture, allowing deeper training.
These models have improved the state-of-the-art in several natural language tasks.

In an effort to disseminate the use of these state-of-the-art methods, this work pre-trained BERT and ALBERT in Portuguese.
Portuguese is a major language in terms of the number of native speakers, but it is still underrepresented in terms of linguistic resources compared to other European languages. 
We believe our work can help mitigate this issue.

\section{Pre-training \label{sec:pre_training}} 
BERT and ALBERT frameworks are designed to be used in two steps: pre-training and fine-tuning.
In the first step, the unsupervised model is pre-trained using a large corpus.
At this stage, the model learns the language features using two training objectives. BERT uses Masked Language Model (MLM) and Next Sentence Prediction (NSP), while ALBERT changes the NSP objective to a Sentence Ordering objective (SO).
In MLM, at random, 15\% of the tokens are replaced by a \texttt{[MASK]} token and the model is supposed to guess which was the best token to be put in its place.
NSP requires that, during pre-training, the input is always composed of two sentences and the model should learn if the second sentence correctly follows the first.
ALBERT's authors showed that the NSP objective is easy, so they changed it to a SO objective to force the model to learn deeper features.
The combination of these objectives forces the model to learn many language features.

\noindent\textbf{Preparation. } 
Although pre-training is an unsupervised task (\ie it just needs a raw document corpus), it requires each training instance to be in a specific format.
The documents are split into sentences and the sentences are tokenized.
BERT uses WordPiece~\cite{wu2016googles}, while ALBERT uses SentencePiece~\cite{kudo-richardson-2018-sentencepiece}.
We used the vocabulary size of 30,000 following the same size used by both BERT and ALBERT when training their single language English model.

%
Portuguese uses diacritics and, while the text can be understood without them, removing them introduces noise as some discriminating features are lost -- \eg the distinction between baby (\textit{bebê}) and s/he drinks (\textit{bebe}) is on the diacritical mark. 
Removing diacritics while pre-training could allow the model to have a better interpretation of informal texts that generally do not use them like tweets or short message service (SMS).
Nevertheless, we decided to keep diacritics and the original casing.
The goal was to maintain discriminating features that could be useful in some tasks. 

Ideally, the pre-trained model should be exposed to texts in the format that it will later be fine-tuned with.
In other words, the same pre-processing steps should be applied both when training and evaluating.
%
%


%
We used 4.8GB of text (992 million tokens) from different kinds of sources.
Because each model has a different vocabulary and tokenizer, the creation of pre-training data must be done for either \nome and \albert.

\noindent\textbf{Corpora. } 
Pre-training is unsupervised and it requires a large corpus. 
To allow for the model to encompass different text styles (formal and informal), Brazilian (BP) and European (EP) variants, we used corpora form a variety of sources. 
\begin{itemize}
    \item \textbf{Wikipedia-PT}\footnote{\url{https://dumps.wikimedia.org/backup-index.html}}, with 8M sentences (after pre-processing) in formal writing, casing, and its contents include all regional variations of Portuguese.  
    
    \item The \textbf{Open Subtitles} corpus\footnote{\url{http://opus.nlpl.eu/OpenSubtitles-v2016.php}} in BP was used as a source of informal language (as it represents spoken language containing slangs and curses). 
    Sentences are short, frequently having fewer than five words.
    
    \item \textbf{News articles} from two corpora:
    ($i$) the CHAVE corpus\footnote{\url{https://www.linguateca.pt/CHAVE/}}, which contains full news articles from the Portuguese Público\footnote{\url{http://www.publico.pt/}} and the Brazilian Folha de São Paulo\footnote{\url{https://www.folha.uol.com.br/}}. 
    Combined, they have a total of 106M tokens; and
    ($ii$) a news corpus from Kaggle\footnote{\url{https://www.kaggle.com/marlesson/news-of-the-site-folhauol}} containing 167K news articles from \textit{Folha de São Paulo}.
 
    \item The \textbf{EuroParl} corpus extracted from the proceedings of the European Parliament\footnote{\url{https://www.europarl.europa.eu/}}. The Portuguese sub-corpus contains about 75M tokens and  it is available at the Open Parallel Corpus site\footnote{\url{http://opus.nlpl.eu/}}. Texts are formally written in EP.
    
    \item \textbf{Research abstracts} from 23K MSc theses and PhD dissertations
    from several areas. The sources were taken from the Brazilian website \emph{Domínio Publico}~\footnote{\url{http://www.dominiopublico.gov.br/}}. This corpus is also made available together with our pre-trained models\footnote{\url{https://github.com/diego-feijo/bertpt/}}. Texts are formally written in BP.
    
\end{itemize}

\noindent\textbf{Parameters. }
During pre-training, we used whole word masking to avoid that a token being masked in the middle. 
For both BERT and ALBERT, the default base configurations were used -- 12 layers, 768 hidden size, and 12 heads of attention.
With this configuration, the model used for \nome has a total of 110M parameters.  As for \albert, there are only 12M parameters.
%

Following the recommended pre-training procedures, \nome was trained for 1M steps and \albert, 175K steps. 
Due to the high number of parameters, BERT takes longer than Albert, and training with longer sequences becomes quite expensive.
As the complexity is quadratic to the length of the sequence, our models were pre-trained with sequences of lengths 128 and 512.
Their training took 33 and 17 hours, respectively, on one cloud TPU v2.

\section{Evaluation \label{sec:experiments}}
In this section, we report on experiments that fine-tune the pre-trained modes \nome and \albert.
The goal was is to compare our trained monolingual models to the provided multilingual and to the current state-of-the-art when applied to several different Natural Language Understanding tasks.
In order to focus on the capacity of the models in extracting features from texts, our models have simple architectures.
They are composed of the standard base model and the pooled output from the last layer is then used as input for a classifier or regression layer.
The new parameters introduced are just the weights from this last layer.
This pooled output is a vector that should be able to capture the desired features to solve the task.
Fine-tuning to each specific task took around five minutes of training using one cloud TPU v2.

Next, we report on our experiments on the following tasks:
($i$) semantic textual similarity, 
($ii$) recognizing textual entailment, 
($iii$) identifying offensive texts, 
($iv$) detecting fake news, 
($v$) categorizing news, 
($vi$) classifying sentiment polarity, and
($vii$) identifying emotions.

In all tasks, \nome and \albert were compared to the Multilingual BERT model released by Google Research, and also to published results for each dataset. 

%

\subsection{Recognizing Textual Entailment (RTE) \label{subsec:assin_recognizing_textual_entailment}} 
The ASSIN dataset is used for semantic textual similarity and recognizing textual entailment. It was first used in a shared task in the PROPOR 2016 Conference\footnote{\url{http://propor2016.di.fc.ul.pt/}}.
It has 10K pairs of sentences, which are equally split between Brazilian (BP) and European Portuguese (EP).
Each language variant is composed of three files: training (2,5K pairs), development (500 pairs), and testing (2K pairs).
The sentences were relatively short. As we said before, each tokenizer split tokens differently. The maximum length in this dataset was 78 tokens.

This task involves determining whether the meaning of one sentence is entailed (can be inferred) from the other sentence.
This is a three-class classification problem that requires assigning a label (None, Entailment, or Paraphrase) to the given pair of sentences.
These classes are highly imbalanced, with the \textit{None} class having three times more instances than \textit{Entailment} and nine times more instances than \textit{Paraphrase}.
The evaluation is done using Accuracy and Macro F1.
Macro F1 evaluates the F-measure for each class independently and then takes the average.
The models were trained using both training and development sets.
Ten-fold stratified cross-validation was applied over the training data.
Table~\ref{tab:assin_10_fold} shows results using just EP, just BP, and a concatenation of both EP+BP training data.
There were no published baselines for the evaluation of only BP.

\begin{table}[tbh]
\caption{RTE results using 10-fold cross-validation} 
\centering
\resizebox{1\linewidth}{!}{%
\begin{tabular}{@{}llrr@{}} \toprule
Data & \multicolumn{1}{c}{Model} & Acc & F1-M \\ \midrule
\multirow{4}{*}{EP} & \citet{rocha2018recognizing}  & 0.83 & 0.73 \\
 & \nome                           & 0.86 & 0.76 \\
 & \albert                         & 0.87 & \textbf{0.80} \\
 & \multi                          & \textbf{0.88} & 0.79 \\ \midrule
\multirow{3}{*}{BP} & \nome        & 0.85 & 0.52 \\
 & \albert                         & 0.85 & 0.51 \\
 & \multi                          & \textbf{0.86} & \textbf{0.57} \\ \midrule
\multirow{4}{*}{EP+BP} & \citet{rocha2018recognizing} & 0.82 & 0.70 \\
 & \nome                        & 0.87 & 0.75 \\
 & \albert                      & 0.88 & 0.79 \\
 & \multi                       & \textbf{0.90} & \textbf{0.80} \\ \bottomrule
\end{tabular}}
\label{tab:assin_10_fold}
\end{table}

For this task, all models were superior to previously reported baselines.  
In this dataset, sentences are well-written, contain punctuation and diacritics. 
The models were not harmed for having a few missing tokens in the vocabulary and so they were able to build very good representations of the sentences.
Also, these short sentences (up to 78 tokens) that do not mix different subjects allow the models to generate more specific representations to be used by the output classifier.
Longer sentences would require representing all subjects covered. Consequently, the vector representation of the sentences would be more diluted.


Next, we applied our fine-tuned model to the entire training set and ran the evaluation over the test set.
The results are on Table~\ref{tab:assin_test}.
The first observation is that the BP setting is difficult for all models.
The EP+BP** in the last rows indicates that the model was trained using both EP and BP training sets, but they were evaluated only using the EP test set.
The results of this evaluation can be compared to the first rows, in which the models were trained and evaluated only on EP test set.
Although it makes sense to think that more training data would help the model generalize, we reached the same conclusion as \cite{FialhoMarquesMartinsCoheurQuaresma2016} that the improvement, if any, in this dataset is meaningless.
We could not find a published baseline for the case where the model was trained using EP+BP and evaluated using both test sets.
Again, as expected, BERT models were superior to previous baselines for all combinations.
Analysing 10-fold evaluation and using only the test set, we can conclude that BERT-based models were superior to previous baselines.

\subsection{Semantic Textual Similarity (STS) \label{subsec:assin_semantic_textual_similarity}} 
The goal of Semantic Textual Similarity is to quantify the degree of similarity between two sentences. 
In the ASSIN dataset, similarity is a number ranging from 1 (no similarity) to 5 (high similarity). 
This is a linear regression problem as the output is a similarity score.
The evaluation measures how far the predicted score is from the ground truth using two metrics:
($i$) Pearson Correlation which measures the correlation with the ground truth, so the higher, the better, and 
($ii$) and the Mean Squared Error which measures the square of the difference between the prediction and the ground truth, so the lower the better.
Following the same procedure of Textual Entailment, we used 3,000 pairs for training.
The results are in Table~\ref{tab:assin_test}. Since STS is related to RTE, again, all models obtained better results than the baselines. Also, the multilingual model was again superior to our pre-trained Portuguese models.

\begin{table*}[tbh]
\caption{RTE and STS evaluated on Test Sets. \label{tab:assin_test}} 
\centering
\begin{tabular}{@{}llrrrr@{}} \toprule
\multirow{2}{*}{Data} & \multirow{2}{*}{Model} & \multicolumn{2}{c}{RTE} & \multicolumn{2}{c}{STS} \\
\cmidrule(lr){3-4} \cmidrule(l){5-6}
&  & Acc & F1-M & Pearson & MSE \\
\midrule
\multirow{4}{*}{EP} & \citet{FialhoMarquesMartinsCoheurQuaresma2016} & 0.84 & 0.73 & 0.74 & 0.60\\
 & \nome                                    & 0.84 & 0.69 & 0.79 & 0.54 \\
 & \albert                                  & 0.88 & 0.78 & 0.80 & 0.47 \\
 & \multi                                   & \textbf{0.89} & \textbf{0.81} & \textbf{0.84} & \textbf{0.43} \\ \midrule
\multirow{4}{*}{BP} & \citet{FialhoMarquesMartinsCoheurQuaresma2016} & 0.86 & 0.64 & 0.73 & 0.36 \\
 & \nome                                    & 0.86 & 0.53 & 0.76 & 0.32 \\
 & \albert                                  & 0.87 & \textbf{0.65} & 0.79 & 0.30 \\ 
 & \multi                                   & \textbf{0.88} & 0.55 & \textbf{0.81} & \textbf{0.28} \\ \midrule
\multirow{3}{*}{EP+BP} & \nome                                 & 0.87 & 0.72 & 0.79 & 0.39 \\
 & \albert                               & 0.89 & 0.76 & 0.78 & 0.39 \\
 & \multi                                & \textbf{0.90} & \textbf{0.82} & \textbf{0.83} & \textbf{0.33} \\ \midrule
\multirow{4}{*}{EP+BP**} & \citet{FialhoMarquesMartinsCoheurQuaresma2016} & 0.83 & 0.72 & - & - \\
 & \nome                               & 0.86 & 0.66 & - & - \\
 & \albert                             & 0.88 & 0.79 & - & - \\
 & \multi                              & \textbf{0.91} & \textbf{0.83} & - & - \\
\bottomrule
\end{tabular}
\end{table*}

\subsection{Offensive Comment Identification\label{subsec:offensive_comments_identification}} 

OffComBR-2 and OffComBR-3~\cite{Pelle2017} are variations of a dataset containing comments that were posted by readers about published news in a Brazilian news portal.
The comments are annotated with \emph{yes} or \emph{no} meaning whether the comment was considered offensive by human judges.
In the OffComBR-2 variation, at least two judges found the comment offensive. In the OffComBR-3 variation, all three judges agreed that the comment was offensive.
Intuitively, OffComBR-3 should be less prone to judge bias, and thus be easier to classify.
This dataset imposes a challenge because the comments are written very informally. They contain slangs, profanities, emoticons, and abbreviations.
There is no punctuation, no diacritics, and casing is almost random.

The length of the comments varies significantly.
In some cases, the offensive part is just one or two tokens and it may be not identified as a vocabulary word.
Also, authors of the offensive comments typically try to obfuscate profane language (\eg by inserting spaces or other symbols among the characters of the offensive word) to make the task of identifying these comments more difficult.
WordPiece and SentencePiece tokenizers are able to represent these misspelled tokens using a sequence of single characters and this is an advantage compared to the traditional vocabulary-based representations.
While a sequence of letters may not be enough for the model to build a good feature representation, in some cases, this may suffice.

The results in Table~\ref{tab:offcombr_10_fold} indicate that this task is hard for BERT models.
It was expected that deep learning models would have a significant advantage over the baseline which uses only shallow methods such as SVM and Naïve Bayes over unigrams.
We believe that the models were not able to achieve a significant advantage because, during pre-training, they were never exposed to offensive language.


\begin{table}[tbh]
\caption{Offensive Comment Identification}
\centering
\resizebox{1\linewidth}{!}{%
\begin{tabular}{@{}llrr@{}} \toprule
Data & Model & Acc & F1-W \\ \midrule
\multirow{4}{*}{BR-2} & \citet{Pelle2017}   & -    & \textbf{0.77} \\
 & \nome                     & \textbf{0.78} & \textbf{0.77} \\
 & \albert                   & 0.76 & 0.76 \\
 & \multi                    & 0.76 & 0.74 \\ \midrule
\multirow{4}{*}{BR-3} & \citet{Pelle2017}  & -  & 0.82 \\
 & \nome                     & \textbf{0.84} & \textbf{0.83} \\
 & \albert                   & \textbf{0.84} & 0.81 \\
 & \multi                    & 0.83 & 0.81 \\
\bottomrule
\end{tabular}}
\label{tab:offcombr_10_fold}
\end{table}

\subsection{Fake News Detection\label{subsec:fakebr_fake_news_detection}} 

The Fake.Br Corpus\footnote{\url{https://github.com/roneysco/Fake.br-Corpus}}~\cite{fakebr:18} contains real examples of fake news written in Brazilian Portuguese. 
Each instance was manually fact-checked and is annotated indicating whether it is fake or true.
There are 7,200 instances evenly split between the classes. 
All sentences are well-written, use punctuation, and diacritics according to the grammar.

We ran the evaluation using 5-fold cross-validation over all training data.
The results are in Table~\ref{tab:fakebr_5_fold}.
The baseline reported achieving up to 0.89 in F-measure when combining several classification models.
BERT models have done an impressive job in this dataset -- they outperformed the baseline and achieved almost perfect results.
Three reasons may explain such good results.
The sentences were well-written, so the model was able to build a good context representation.
Despite the maximum length of the texts being more than 512, the majority of the texts have similar lengths. We believe that this similarity allowed the model to better detect any variations that indicate that the news was true or fake.
Finally, we attribute the good performance mostly to the fact that this dataset is reasonably large when compared to the others evaluated here. The larger volume of training data allowed the model to learn enough features to distinguish fake from true.


\begin{table}[tbh]
\caption{Fake News Detection}
\centering
\begin{tabular}{@{}lrr@{}} \toprule
Model & Acc & F1-W \\ \midrule
\citet{fakebr:18}             & 0.89 & 0.89 \\
\nome                         & \textbf{0.98} & \textbf{0.98} \\
\albert                       & \textbf{0.98} & \textbf{0.98} \\
\multi                        & \textbf{0.98} & \textbf{0.98} \\
\bottomrule
\end{tabular}
\label{tab:fakebr_5_fold} 
\end{table}

\subsection{Sentiment Polarity Classification on Tweets \label{subsec:emocoesbr_sentiment_analysis}} 

In \cite{araujo2016@sac} the performance of several models for sentiment polarity classification were evaluated.
The best reported macro F1 with more than 90\% of messages classified was 0.71.
We use this score as the baseline for Portuguese, although the authors did not report how it was obtained.

There are three possible sentiments: negative, neutral, or positive.
The writing style is very informal and many out of vocabulary were used. Expressions like ``goooood daaaay'', ``loool'', ``kkk'', ``=D'' are frequent.
So, if the model only relies on known words, this task may be difficult.
In this situation, the fact that the model uses subword units to tokenize the text may help to at least try to build an interpretation using the sequence of units.
Also, some examples expose the sentiment through emoticons.
So, correctly interpreting these symbols can lead to the associated sentiment.

In order keep the results comparable, we use only the positive and negative samples.
For the results shown in Table~\ref{tab:sentiment_5_fold}, we did a 5-fold cross-validation over all data.
The fact that this corpus has several emoticons and out-of-vocabulary expressions makes it hard for the models that were not trained using a similar vocabulary.


\begin{table}[tbh]
\caption{Sentiment Polarity Classification}
\centering
\begin{tabular}{@{}lrr@{}} \toprule
Model & Acc & F1-W \\ \midrule
\citet{araujo2016@sac}        & -    & 0.71 \\
\nome                         & 0.77 & 0.76 \\
\albert                       & \textbf{0.79} & \textbf{0.78} \\
\multi                        & 0.71 & 0.70 \\
\bottomrule
\end{tabular}
\label{tab:sentiment_5_fold} 
\end{table}

\subsection{News Category Classification} 
\label{subsec:folhauol_news_category_classification}
The Folha UOL News Dataset\footnote{\url{https://www.kaggle.com/marlesson/news-of-the-site-folhauol}} contains 167,053 news from  \emph{Folha de São Paulo}\footnote{\url{https://www.folha.uol.com.br/}}, a Brazilian newspaper. It contains the headlines, complete articles, and their category (opinion, daily life, sports, culture, markets, world, and politics). 

There is a public ranking of this classification task on Kaggle\footnotemark.
The best result so far used 15\% of the data as validation set and achieved 87.39\% of validation accuracy.
\footnotetext{\url{www.kaggle.com/marlesson/news-of-the-site-folhauol/kernels}}
%
%
To keep results as comparable as possible, we followed the ranking leader and randomly split 15\% to be used as our test set. As this could be an arbitrary split, we also ran the evaluation using stratified 5-fold cross-validation.
The results are shown in Table~\ref{tab:folhauol_5_fold}. All models behaved almost identically, with a slight advantage for the multilingual model.

\begin{table}[tbh]
\caption{Folha Uol (7-classes)}
\centering
\begin{tabular}{@{}llrr@{}} \toprule
Model & Mode & Acc & F1-W \\ \midrule
Kaggle Baseline\footnotemark[\value{footnote}] & Split 15\% & 0.87 & - \\
\nome                & Split 15\% & 0.94 & \textbf{0.94} \\
\albert              & Split 15\% & 0.93 & 0.93 \\
\multi               & Split 15\% &\textbf{0.94} & \textbf{0.94} \\ \midrule
\nome                & 5-fold  & \textbf{0.94} & \textbf{0.94} \\
\albert              & 5-fold  & 0.93 & 0.93 \\
\multi               & 5-fold  &\textbf{0.94} & \textbf{0.94} \\ \bottomrule
\end{tabular}
\label{tab:folhauol_5_fold}
\end{table}


Another experiment of news category classification was done using Público Dataset. Público\footnote{\url{https://www.publico.pt/}} is a Portuguese Newspaper. The dataset containing its news is distributed as part of CHAVE corpus\footnote{\url{https://www.linguateca.pt/CHAVE/}}.
Each news text belongs to one out of nine categories: Science, Culture, Sports, General, Economy, Local, World, National, and Society.
The task is a multiclass classification with a single label.
For this evaluation, we used a 5-fold cross-validation.
The best reported baseline achieved 0.84 in F-measure when combining several classification models.

The news texts are well written, contain punctuation and diacritics.
The major problem here is length diversity -- while the mean length is 728 tokens, the standard deviation is 628.
Both BERT and ALBERT models were not able to handle such long texts due to memory limitations.
We imposed a limit of using only the first 512 tokens of the text.

Table~\ref{tab:publico_5_fold} has the results for this task.
Without any task-specific architecture, the performance in this dataset was almost identical to the current state-of-the-art in this task.

\begin{table}[tbh]
\caption{Público News (9-classes) using 5-fold}
\centering
\begin{tabular}{@{}lrr@{}} \toprule
Model & Acc & F1-M \\ \midrule
\citet{teresa2008using}   & \textbf{0.84} & \textbf{0.84} \\
\nome                & \textbf{0.84} & \textbf{0.84} \\
\albert              & 0.82 & 0.82 \\
\multi               & \textbf{0.84} & \textbf{0.84} \\ \bottomrule
\end{tabular}
\label{tab:publico_5_fold}
\end{table}

\subsection{Emotion Classification \label{subsec:brnews_sentiment_analysis}} 
The BRNews Dataset~\cite{martinazzo2012} contains news extracted from major Brazilian newspapers.
Each of the 1,002 instances was manually annotated for one within six possible emotions: joy, surprise, anger, disgust, fear, and sadness.

The texts from the news are short and well-written. They use punctuation and diacritics.
The length in tokens varies from 21 to 62 tokens, which allows the model to capture most of the meaning of the sentences.
The most represented class has up to ten times the frequency of the two least represented.

We ran two kinds of experiments.
In the first, we employed a standard multiclass classifier.
Unfortunately, we do not have any baseline results, so we just report the results for the BERT-based models.
In the second experimental setting, we followed the same methodology as~\cite{Becker:2017:MEC:3063600.3063706} and transformed the problem into six binary classifiers (one for each sentiment).
For each one of these binary models, the accuracy and weighted F1 score were calculated. Then, the mean result was taken.

Table~\ref{tab:brnews_10_fold} shows the results for 10-fold cross-validation.
The baseline for the binary version reaches up to 0.84 of F-measure. In this experiment, \nome has a result similar to the baseline, but it was not able to overcome it.

\begin{table}[htb]
\caption{Emotion Classification}
\centering
\resizebox{1\linewidth}{!}{%
\begin{tabular}{@{}lrrrr@{}} \toprule
\multirow{2}{*}{Model} & \multicolumn{2}{c}{6 classes} & \multicolumn{2}{c}{Binary} \\
\cmidrule(lr){2-3} \cmidrule(lr){4-5}
 & Acc & F1-W & Acc & F1-W \\ \midrule
\citet{Becker:2017:MEC:3063600.3063706} & - & - & -  & \textbf{0.84} \\
\nome                              & \textbf{0.51} & \textbf{0.47} & \textbf{0.84} & 0.83 \\
\albert                            & 0.41 & 0.28 & \textbf{0.84} & 0.81 \\
\multi                             & 0.49 & 0.46 & \textbf{0.84} & 0.80 \\
\bottomrule
\end{tabular}}
\label{tab:brnews_10_fold}
\end{table}

Inspecting some misclassifications, we see some unusual annotations. For example, ``President Lula will be in Sergipe this Friday for inauguration of works: Lula will visit three cities. The ally Governor will be joining him.'' This instance was annotated as ``fear'', but \nome predicted ``joy''.
So, we think that the fact that the models were unable to beat the established baseline is more due to dataset characteristics than the ability of the model to interpret its meaning.


\section{Discussion}

The different tasks and settings under which \nome and \albert were compared to Multilingual BERT and published baselines yield 38 possible pairwise comparisons.
We analyzed in how many times the performance of each model was better than, worse than or equivalent to the performance of the others. If the proportional difference in scores was below 5\%, then the two models were considered equivalent.
These results are shown in Table~\ref{tab:comparison}.
We can see that no single method is always better than the others.
The baselines have fewer wins, which means that BERT-based models tend to yield better results. 
\nome and \albert achieved equivalent performances most of the time (28 out of 38).
\albert has an advantage (6 wins versus 4 losses), and since its training is less expensive, then it would be the model of choice.

Overall, Multilingual BERT has more wins in comparison to other alternatives. Although the differences lie most of the time within the 5\% range, it achieved better results in more tasks.
We attribute this superiority to the use of more languages. It is possible that the presence of languages that are similar to Portuguese (such as Spanish, which has twice the volume of texts of Portuguese) was able to produce richer semantic representations.

\begin{table}[tbh]
\caption{Comparison of the 3 models across all tasks} \label{tab:comparison}
\centering
\resizebox{1\linewidth}{!}{%
\begin{tabular}{@{}l rrr rrr rrr@{}} \toprule
\multirow{2}{*}{Model} & \multicolumn{3}{c}{BertPT}                                                                   & \multicolumn{3}{c}{AlbertPT}                                                                 & \multicolumn{3}{c}{Multilingual} \\
                  & \multicolumn{1}{c}{\textgreater{}} & \multicolumn{1}{c}{=} & \multicolumn{1}{c}{\textless{}} & \multicolumn{1}{c}{\textgreater{}} & \multicolumn{1}{c}{=} & \multicolumn{1}{c}{\textless{}} & \multicolumn{1}{c}{\textgreater{}} & \multicolumn{1}{c}{=} & \multicolumn{1}{c}{\textless{}} \\ \midrule
Baseline          & 3                                  & 12                    & 8                               & 1                                  & 8                     & 14                              & 1                                  & 13                    & 9                               \\
BertPT            &                                    &                       &                                 & 4                                  & 28                    & 6                               & 2                                  & 24                    & 12                              \\
AlbertPT          &                                    &                       &                                 &                                    &                       &                                 & 1                                  & 30                    & 7      \\
\bottomrule
\end{tabular}}
\end{table}

The limitation on the size of the input (which could have a higher impact on the news categorization task) ended up proving not to be an issue -- classification accuracy was very high in both datasets. Classification errors happened in cases in which the distinction of the categories was quite fuzzy (\eg daily life and  culture).

Multilingual BERT was not trained on informal texts, which were abundant in two of our tasks (offensive comment detection and sentiment polarity classification). As a result, the results of \nome and \albert were slightly superior in these tasks, showing that having more text styles on the training set is beneficial.


\balance

\section{Conclusion\label{conclusion}}

This work presented \nome and \albert, pre-trained language models for Portuguese.
We described the pre-training procedure along with the resources we used.
Then, the models were evaluated in seven natural language understanding tasks. The results were compared with the official Multilingual BERT and existing baselines for each task.
The good results achieved by \albert and the Multilingual BERT, indicate that having an Albert-based multilingual language model would be a very useful resource.
As future work, we intend to pre-train a multilingual model using languages with good availability of training data that are similar to Portuguese such as Spanish, French, and Italian.

\bibliography{references.bib}
\bibliographystyle{acl_natbib}

\end{document}